\documentclass[letterpaper, 10 pt, conference]{ieeeconf}  

\IEEEoverridecommandlockouts                              

\usepackage{graphicx}
\usepackage{amsfonts,amssymb}
\usepackage{multirow}
\usepackage{hyperref}
\usepackage[ruled]{algorithm2e}
\usepackage{algorithmic}
\usepackage{pdfpages}
\usepackage{subfigure}
\usepackage{tabularx}
\usepackage{float}
\usepackage{setspace,graphicx,booktabs,color,url,mathtools,soul}

\title{\LARGE \bf
Deep Reinforcement Learning for Solving Management Problems: Towards A Large Management Model 
}

\author{Jinyang Jiang$^{1*}$, Xiaotian Liu$^{2*}$, Tao Ren$^{1*}$, Qinghao Wang$^{3*}$, Yi Zheng$^{1*}$,\\ Yufu Du$^{4}$, Yijie Peng$^{1}$ and Cheng Zhang$^{4}$   
\thanks{*These authors contributed equally to this work.}
\thanks{$^{1}$Jinyang Jiang, Tao Ren, Yi Zheng, and Yijie Peng are with Wuhan Institute for Artificial Intelligence, Guanghua School of Management, Peking University and Xiangjiang Laboratory, China
        {\tt\small \{jinyang.jiang, rtkenny, yizheng\}@stu.pku.edu.cn, pengyijie@pku.edu.cn}}%
\thanks{$^{2}$Xiaotian Liu is with H. Milton Stewart School of Industrial and Systems Engineering,
Georgia Institute of Technology, Atlanta, USA
        {\tt\small xliu939@gatech.edu}}%
\thanks{$^{3}$Qinghao Wang is with Institute for Artificial Intelligence, Peking University, Beijing 100871, China
        {\tt\small qinghw@pku.edu.cn}}%
\thanks{$^{4}$Yufu Du and Cheng Zhang are with Random Seed Inc. China
        {\tt\small 8870595@qq.com, chengzhang@gsm.pku.edu.cn}}%
}

\begin{document}
\maketitle
\thispagestyle{empty}
\pagestyle{empty}

\begin{abstract}
We introduce a deep reinforcement learning (DRL) approach for solving management problems including inventory management, dynamic pricing, and recommendation. This DRL approach has the potential to lead to a large management model based on certain transformer neural network structures, resulting in an artificial general intelligence paradigm for various management tasks. Traditional methods have limitations for solving complex real-world problems, and we demonstrate how DRL can surpass existing heuristic approaches for solving management tasks. We aim to solve the problems in a unified framework, considering the interconnections between different tasks. Central to our methodology is the development of a foundational decision model coordinating decisions across the different domains through generative decision-making. Our experimental results affirm the effectiveness of our DRL-based framework in complex and dynamic business environments. This work opens new pathways for the application of DRL in management problems, highlighting its potential to revolutionize traditional business management.

\end{abstract}

\section{INTRODUCTION}


Effective coordination across inventory management, pricing, and recommendations is crucial in the business. For example, In today's rapidly evolving fashion industry, the process, from the initial design to production and final marketing, faces complex management challenges at every step: How do companies ensure supply chain flexibility and cost efficiency under unstable supplies? How do managers set the price according to the inventory and the demand? How do companies flexibly adjust the recommendation policy based on current inventory conditions?

In the fashion industry, where the seasonality of demand, the rapid change of trends, and the uncertainty of supply chains make management complicated, finding more effective solutions becomes significant. The fluctuation of supply and demand poses challenges for inventory management. The seasonal changes in demand require a good pricing policy. The ever-changing fashion trends and customers' preferences make the precise recommendation difficult. An efficient management solution should jointly optimize inventory, pricing, and recommendation. However, most prior studies have focused on using analytical and heuristic methods to solve these problems independently \cite{clark1960optimal}\cite{feng2014dynamic}\cite{demirezen2016optimization}. The analytical and heuristic solutions often fail in more complex real-world business scenarios, i.e. fashion industry.

To address this challenge, we propose a universal framework based on Deep Reinforcement Learning (DRL) to comprehensively solve management problems in inventory management, pricing, and recommendation systems. 

Recent advancements in DRL have shown promising applications in industrial scenarios, notably in inventory management \cite{liu:2023}. We suggest that single and multi echelons inventory problems can be solved in a unified DRL framework. We extend DRL's application to dynamic pricing \cite{elmaghraby2003dynamic} and inventory replenishment, demonstrating our algorithm's superior performance over classical methods \cite{feng2020integrating}\cite{lei2018joint} by mitigating demand mismatches and maximizing profits. Furthermore, we integrate DRL with recommendation systems, addressing the critical aspect of recommendation intensity tailored to inventory management, thereby showcasing our framework's enhanced efficiency and effectiveness in managing complex business operations.  

Inspired by the Large Language Models, we propose to build the foundational decision model for inventory management, pricing, and recommendation systems. Utilizing autoregressive architecture \cite{decisiontran}\cite{trajectory}\cite{Bommasani2021}, the model can learn and optimize past trajectories, yielding favorable outcomes. The foundation model is a prototype of the Large Management Model (LMM). We believe that through large-scale pre-training from historical data, LMM will possess universal decision-making capabilities in management and lead us one step closer to artificial general intelligence (AGI).

\section{RELATED WORKS}

\subsection{Inventory Management}
The most classic inventory management literature considers a single-sourcing backlogged model with stationary customer demand and a deterministic lead time. The $(s, S)$ policy makes an order to elevate the inventory level to S as soon as it drops below $s$. $s$ and $S$ are two exogenous parameters needed to be determined in the $(s, S)$ policy \cite{clark1960optimal}. Given non-stationary customer demand and stochastic lead time, an analytical optimal policy is intractable, prompting Qi et al. \cite{qi2023practical} to develop near-optimal ordering policies using a one-step, end-to-end deep learning framework.


\subsection{Dynamic Pricing and Replenishment}
The existing literature on dynamic pricing and inventory control often neglects real market conditions, such as market competition and interdependencies between products, when modeling demand \cite{federgruen1999combined}. Additionally, it relies on strong assumptions regarding the demand information for the products \cite{feng2014dynamic}. The application of their strategies to practical scenarios encounters many challenges owing to these unrealistic assumptions. 

Few studies have explored the application of DRL to the joint dynamic pricing and replenishment problem \cite{liu4395571data}. The separation of decision-making processes indicates their study doesn't fully achieve joint optimization of pricing and replenishment.

\subsection{Recommendation System}
The recommendation system is at the intersection of marketing and information systems \cite{ricci2021recommender}. It recommends content to customers and influences their near-term purchasing behaviors. \cite{ghoshal2020dilemma} uses game theories to figure out how companies compete with others to enhance the business output. The complexity of integration between recommendation systems and operational management has hindered its implementation \cite{tang2010review}. The only relevant study utilizes mixed integer programming to adjust movie rental recommendation strategies based on the inventory, but it only optimizes the recommendations unilaterally \cite{demirezen2016optimization}. 


\section{PRELIMINARY}
To solve the management problems with DRL, we first need to formulate the problems as Markov Decision Processes (MDP). Specifically, an MDP is defined by five elements $\{\mathbb{S}, \mathbb{A}, \mathcal{R}, \mathcal{T}, \gamma\}$, where $\mathbb{S}$ represents the state space, $\mathbb{A}$ represents the action space, $\mathcal{R}:\mathbb{S}\times\mathbb{A}\rightarrow\mathbb{R}$ represents the reward function, $\mathcal{T}:\mathbb{S}\times\mathbb{A}\rightarrow\mathbb{S}$ represents the state transition function, and $\gamma \in (0, 1]$ is the discount factor. Then applying state-of-art DRL algorithms, such as PPO \cite{Schulman:2017} and A3C \cite{Mnih:2016}, the optimal policy of the process can be found.


\section{INVENTORY MANAGEMENT}
\subsection{Single-Echelon Models}
The single-echelon inventory models consider inventory decisions for one actor in one stage of a supply chain. These problems can be naturally formulated as standard MDPs where DRL applies. Under the context of single-echelon inventory management, the state $s_t\in\mathbb{S}$ at time step $t$ typically contains basic managerial information such as on-hand inventory, outstanding orders, etc. The action $s_t\in\mathbb{A}$ is mapped into the ordering quantity at period $t$. The reward $R(s_t, a_t)$ is defined by the one-period profits or the opposite of one-period costs. The inventory policy $\pi:
 \mathbb{S}\rightarrow\mathbb{A}$ parameterized by Neural Networks (NNs) takes state $s_t$ at each period $t$ as input and outputs the action $a_t$. The objective of DRL is to tune the parameterized policy $\pi$ such that the expected long-term rewards are maximized, i.e.,
 $$
 \max_{\pi}\mathbb{E}_{a_{0:\infty}\sim\pi,s_{0:\infty}\sim\mathcal{T}}\bigg[\sum_{t=0}^\infty\gamma^tR_t(s_t, a_t)\bigg].
 $$
\begin{table*}[!t]
    \centering
    \caption{Results of applying DRL methods on single-echelon inventory problems.}
    \label{tab:res single-echelon}
    \small
    \begin{tabular}{c|c>{\centering\arraybackslash}m{4cm}m{10cm}}\toprule
\midrule
\textbf{Paper}              & \textbf{Method} & \textbf{Inventory Settings}                                                                      & \multicolumn{1}{c}{\textbf{Main Results}}         \\ \midrule
\multirow{4}{*}{\cite{Gijsbrechts:2021}} & A3C               & Lost sales                                                                              & A3C beats base stock policy, constant order policy, and Myopic 1-period policy, but is inferior to Myopic 2-period policy and capped base stock policy. \\ \cmidrule{2-4}
                   & A3C               & Dual sourcing                                                                           & A3C achieves comparable performance with single index policy and tailored base surge policy but is slightly inferior to optimal capped dual index policy.   \\ \midrule
\cite{liu:2023}                  & PPO    & Periodic review with uneven review intervals & PPO approximately converges to the optimal base stock policy.                                                                                               \\ \midrule
\bottomrule
\end{tabular}
\end{table*}


The single-echelon inventory problems have been intensively studied in the classic inventory literature and numerous optimal or near-optimal heuristic policies have been established \cite{snyder:2019}. Recent results show, under multiple single-echelon inventory settings, that DRL can achieve comparable performance to these optimal or near-optimal heuristics. We summarize several representative results in Tab. \ref{tab:res single-echelon}.

\subsection{Multi-Echelon Models}
The multi-echelon inventory models consider the ordering decisions for multi-actors in the supply chain. These problems complicate the single-echelon models by involving the coordination and interaction among multiple actors in the supply chain. To apply DRL, the multi-echelon inventory problems are typically formulated as Markov Games. Under a fully cooperative setting, each actor has the same objective, i.e., $\mathcal{R}^1(\cdot)=\cdots=\mathcal{R}^M(\cdot)=\mathcal{R}(\cdot)$, and the objective of DRL is to find the optimal joint policy that maximizes the shared expected long-term rewards, i.e.,

\begin{equation}
\begin{aligned} 
\max_{\pi^1, \ldots,\pi^M}\mathbb{E}_{a^1_{0:\infty}\sim\pi^1,\ldots,a^M_{0:\infty}\sim\pi^M, s_{0:\infty}\sim\mathcal{T}} \\
\bigg[\sum_{t=0}^\infty\mathcal{R}(s_t, a^1_t, \ldots, a^M_t)\bigg].
\end{aligned}
\end{equation}

Under the non-fully cooperative setting, objective of each actor $i\in\{1,\ldots,M\}$ is to find its individual optimal policy that maximizes its own expected long-term rewards given the policy $\tilde{\pi}^1,\ldots,\tilde{\pi}^{i-1}, \tilde{\pi}^{i+1}, \ldots,\tilde{\pi}^{M}$ of other actors, i.e.,


\begin{equation}
\max_{\pi^i}\mathbb{E}_{\boldsymbol{a}_{0:\infty}\sim\boldsymbol{\pi}, s_{0:\infty}\sim\mathcal{T}}\bigg[\sum_{t=0}^\infty\mathcal{R}^i(s_t, a^1_t, \ldots, a^M_t)\bigg],
\end{equation}
where $\boldsymbol{a}_{0:\infty}=(a^1_{0:\infty},\ldots,a^M_{0:\infty})$ and $\boldsymbol{\pi}=(\tilde{\pi}^1,\ldots,\tilde{\pi}^{i-1}, \pi^{i}, \tilde{\pi}^{i+1},\ldots,\tilde{\pi}^{M})$. Developing effective policies for multi-echelon inventory models faces the following challenges.
\begin{itemize}
    \item The joint policy space is large under the fully cooperative setting.
    \item The system dynamics are too complicate to develop analytical models or heuristic policies.
    \item The interaction of multiple actors under the non-fully cooperative settings makes it hard to characterize the equilibrium joint policy.
\end{itemize}

For the multi-echelon inventory models, recent results have shown the superiority of DRL policies in terms of performance and generalization ability on different supply structures. Some representative results are summarized in Tab. \ref{tab:res multi-echelon}. These results well demonstrate the potential of DRL method in solving highly complex real-life problems. One of the main advantages of DRL method is that it has little requirement on the problem settings, which means it can be easily applied under various real-life scenarios. Supported by these existing results, it is reasonable to expect a large DRL model
to solve highly complex inventory problems and achieve superb performance. 
 
\begin{table*}[!t]
\centering
\caption{Results of applying DRL methods on multi-echelon inventory problems.}
\label{tab:res multi-echelon}
\small
\begin{tabular}{c|c>{\centering\arraybackslash}m{5cm}m{9cm}}\toprule
\midrule
\textbf{Paper}              & \textbf{Method} & \textbf{Inventory Settings}                                                                                        & \multicolumn{1}{c}{\textbf{Main Findings}}                                                                                                                                  \\ \midrule
\cite{Oroojlooyjadid:2022}                  & DQN    & 
Beer game in serial supply chain                         & DQN exceeds the base stock policy when other supply chain actors make realistic ordering decisions.                                        \\ \midrule
\multirow{3}{*}{\cite{liu:2023b}} & HAPPO  & Beer game in serial supply chain with non-stationary demands & HAPPO surpasses the non-stationary base stock policy without fixed costs and the non-stationary ($s$, $S$) policy with fixed costs. \\ \cmidrule{2-4}
& HAPPO  & A supply chain network where each echelon has two actors    & HAPPO outperforms the capped dual index, dual index, and tailored base surge policies adapted to the problem.                                              \\ \midrule
\cite{Gijsbrechts:2021}                  & A3C    & One-warehouse multi-retailers                                  & A3C outperforms the base stock policy with constant base-stock levels.                                                                                     \\ \midrule
\cite{liu:2023}                  & PPO    & One-warehouse multi-retailers with dual sourcing and multi-products     & PPO outperforms three combined heuristic policies.                                                                                  \\ \midrule
\bottomrule
\end{tabular}
\end{table*}

\section{DYNAMIC PRICING AND REPLENISHMENT}
In this part, we formulate the dynamic pricing and replenishment under competition problem as an MDP, followed by an illustration of the application of DRL and a performance comparison between the baseline policies and DRL.

\subsection{Dynamics of the Dynamic Pricing and Replenishment Model}\label{AA}
Consider a $T$-period problem. At the beginning of each period $t$, $t = 1,...,T$, the replenishment quantity $q_t$ and price $p_t$ for a product must be decided. We infer overall demand by modeling the probability of each individual consumer's choice based on real-world data \cite{schlosser2019dynamic}. Given our offer price $p$, the competitor’s price $o$, and the current reference price $r$, the demand for a product in period $t$ is denoted by
\begin{equation} \label{eq:2}
    D_t  \sim \text{Pois}\Big( \eta \Delta \cdot e^{\bold{x}(p,o,r)^{'}\bold{\beta}} \text{/} (1+e^{\bold{x}(p,o,r)^{'}\bold{\beta}}) \Big).
\end{equation}
where $\Delta \in [0,1]$ is a time span of length, $\eta$ is a scaling factor, $\bold{x}(p,o,r)$ represents regressors relating to sales and $\bold{\beta}=(\beta_1,...,\beta_6)$ represents the corresponding parameter vector. For more details, please refer to \cite{schlosser2019dynamic}. 
The dynamics of the evolution of inventory can be expressed by \par
\begin{equation} \label{eq:3}
    I_t = [I_{t-1}+q_{t-{L}}-d_t]^+,
\end{equation}
where $d_t$ is the realization of $D_t$, $L$ is the length of lead time.
The actual sales $A_t$ is given by
\begin{equation}\label{eq:4}
    A_t = \min \{d_t,I_{t-1}+q_{t-{L}}\}.
\end{equation}
The backlog $B_t$ is 
\begin{equation}\label{eq:5}
    B_t = \max \{d_t-(I_{t-1}+q_{t-{L}}),0\}.
\end{equation}
The sales profit $r_t$ is
\begin{equation}\label{eq:6}
    r_t = p_t A_t - h I_t - b B_t - c q_t,
\end{equation}
where $h,b,c$ are unit holding, backlogged, and ordering costs, respectively.

To apply DRL algorithms, we formulate the dynamic pricing and replenishment as an MDP, where the actions are defined as $a_t=\{p_t,q_t\}$. We define product's state in period $t$ as $s_t = \{I_{t-1},B_{t-1},d_{t-1},q_{t-{L}},...,q_{t-1}\}$. Eq. (\ref{eq:3})-(\ref{eq:5}) describe the dynamics. The period reward is defined as Eq. (\ref{eq:6}) to maximize the total sales profit. We parameterize our pricing and replenishment policy as a conditional distribution over the action space given the state by neural networks and adopt PPO to solve the MDP.

\subsection{Baseline Policies and Experimental Results}
We implement three classic heuristic policies as baselines to evaluate the performance of PPO. 
\begin{itemize}
\item The Base-stock List-price (BSLP) policy \cite{thowsen1975dynamic}, which is optimal for the additive demand function by assuming that the random noise follows a Pólya frequency function of order 2 (PF2) distribution. 
\item The (s,S,p) policy \cite{chen2004coordinating}, which has been shown to be optimal when the ordering cost comprises a fixed component and a variable cost proportional to the ordered quantity. In this case, both inventory holding and shortage costs are convex functions of the inventory level. 
\item The 'Myopic' policy \cite{bernstein2016simple}, which involves a myopic pricing policy that determines each period’s price $p_t$ as a function of the beginning inventory level $x_t$ at time $t$. Additionally, it employs a base-stock policy for inventory replenishment and has shown excellent performance in numerical studies.
\end{itemize}

\begin{figure}[htbp]
\centerline{\includegraphics[width=\columnwidth]{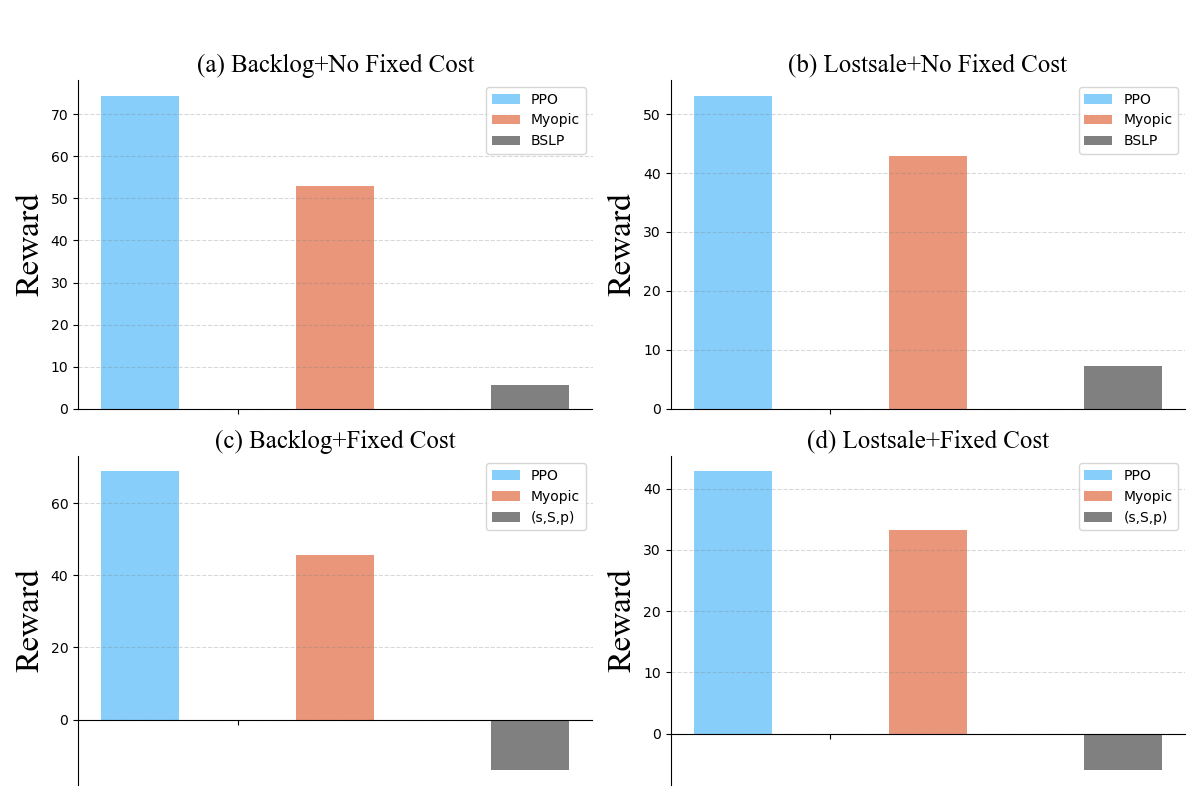}}
\caption{Comparison of reward under different scenarios.}
\label{fig:Comparison}
\end{figure}

We evaluate PPO's performance on dynamic pricing and replenishment in competition through four experiments: (a) backlogged demand with no fixed ordering costs; (b) lost demand with no fixed ordering costs; (c) backlogged demand with fixed ordering costs; (d) lost demand with fixed ordering costs. With demands generated by Eq. (\ref{eq:2}), comparisons of reward in Fig. \ref{fig:Comparison} show that PPO achieves the highest profits in all cases.

\section{INVENTORY MANAGEMENT WITH RECOMMENDATION SYSTEM}


In this part, we formulate the coordination of the inventory management and recommendation system as an MDP. With DRL as a powerful solver, it becomes feasible to jointly consider the two problems. Numerical results are carried out to demonstrate the advantage of coordination via DRL.  

\subsection{Coordination between Recommending and Replenishment}

We use a recommendation system including $M$ customers and $N$ products. The recommending workflow adjusts the exposure situations of products to influence customers' ratings among them. We abstract it as determining the recommending intensity $\alpha_t^{i,j}$ of product $i$ to customer $j$ at period $t$, where $\alpha_t^{i,j}\in[0,1]$ and $\sum_{j=1}^M \alpha_t^{i,j} = 1$. The rating of customer $j$ for product $i$ is denoted as $R^{i,j}_{t}$, and the customer originally holds a rating $R^{i,j}_{-1}$. We model the advertisement phenomenon of recommendation as 
\begin{align}
R^{i,j}_t = R^{i,j}_{-1} + (R_{\max}-R^{i,j}_{-1}) \alpha_t^{i,j} E, \label{eq: recommendation_rate}
\end{align}
where $E$ is the efficiency coefficient, $(R_{\max}-R^{i,j}_{-1})$ is the maximum possible improvement in the rating, indicating a decreasing marginal. The relative rating affects purchasing behaviors \cite{lilien1992marketing}, which determines the external demand $d_t^i$ of the inventory system. We assume that the individual demand $d_t^{i,j}$ is assigned following the relative rating, i.e.,
\begin{align}
d_t^i = \sum_{j=1}^M d_t^{i,j},\quad d_t^{i,j}\sim B(c^j, \gamma_t^{i,j}),
\end{align}
where $d_{t}^{i,j}$ follows a binomial distribution, $c^j$ denotes the purchasing capacity of customer $j$, and $\gamma_t^{i,j}$ is the probabilities over all products given by
\begin{align}
\gamma_t^{i,j} = \exp\{R^{i,j}_t\}\big/\sum_{k=1}^N\exp\{R^{k,j}_t\},
\end{align}
where a higher rating implies the customer is more likely to purchase the product.

We consider a standard $N$-product single-echelon inventory system with lost sales during $T$ periods, where the unmet demand will disappear immediately. We use $S_t^i$, $O_t^i$, and $I_t^i$ to denote the sales quantity, lost sales, and on-hand inventory of product $i$ at the end of period $t$. The replenishment quantity of product $i$ is denoted as $q_t^i$, which is the decision variable in inventory management. The status of the inventory system is given by
\begin{align}
S_t^i&=d_t^{i}-\big[d_t^{i}-I_{t-1}^i-q_{t-L^i}^{i}\big]^{+},\\
O_t^i&=\big[d_t^{i}-S_t^i\big]^{+}, \\
I_t^i&=\big[I_{t-1}^i+q_{t-L^i}^{i}-S_t^i\big]^{+}, \label{eq: inventory_on_hand}
\end{align}
where $[x]^+:= \max\{x,0\}$, and $L^i$ is the duration that product $i$ is in transit, namely the lead time.
The profit of each product at period $t$ is calculated by
\begin{align}
    F_t^i=p^i_{\text{out}} S_t^i-p^{i}_{\text{in}} q_t^{i}-h^i I_t^i-b^i O_t^i,
\end{align}
where $p^i_{\text{out}}$, $p^{i}_{\text{in}}$, $h^i$ and $b^i$ are the unit price for sale and reordering, unit holding cost, and unit penalty for lost sales of product $i$, respectively.

To apply DRL algorithms, we formulate the coordination of inventory management and recommendation system as an MDP, where the actions are defined as $a_t=\{\{q_t^i\}_{i}, \{\{\alpha_t^{i,j}\}_{j}\}_{i}\}$; the states includes the minimal information in past $L=\max_i L_i$ periods required to produce reasonable results, i.e., $s_t=\{\{I_{t-1}^i\}_{i}, \{\{q^i_{t-t'}\}_{t'}\}_{i}\}$; Eq. (\ref{eq: recommendation_rate})-(\ref{eq: inventory_on_hand}) describe the dynamics; and the period reward is computed by $r_t = \sum_{i=1}^N F_t^i$ to maximize the total inventory profit. We parameterize the decision-making policy as a conditional distribution over the action space given the state by a neural network or a cluster of neural networks and adopt policy-based DRL algorithms, e.g., PPO, to solve the MDP.

\subsection{Experimental Results}
\begin{figure}[h]
\centering 
\subfigure[Lost sales.]{
\label{fig:lost_sale_imrs}
\includegraphics[trim=0cm 0.5cm 0cm 0cm, width=0.45\linewidth]{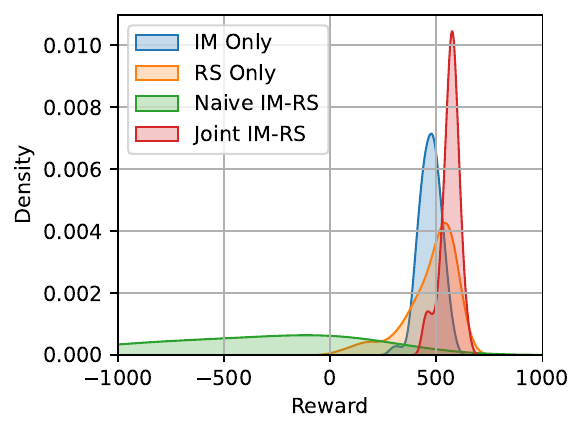}}
\subfigure[Backlog.]{
\label{fig:backlog_imrs}
\includegraphics[trim=0cm 0.5cm 0cm 0cm, width=0.45\linewidth]{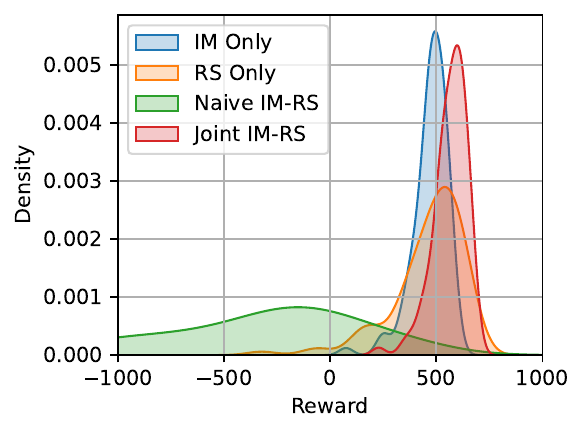}}
\caption{Comparison of reward KDEs under different scenarios.}
\label{fig:imrs}
\end{figure}

We conduct experiments in the lost sale and backlog scenario to demonstrate the advantage of coordination. In each scenario, we train three agents, with inputs being the global state, and outputs being respectively joint actions (Joint IM-RS), actions with recommending decisions replaced by random choices (IM-Only), and actions with replenishment decisions replaced by random choices (RS-Only). We also test the integration of trained agents from IM-Only and RS-Only by using each of them to output part of actions (Naive IM-RS). The kernel density estimation (KDE) plots of episode rewards are shown in Fig. \ref{fig:imrs}. Joint IM-RS offers more concentrated and higher rewards, suggesting coordination between inventory management and recommendation systems is profitable. Naive IM-RS results show that independent operations can worsen outcomes due to demand and supply mismatches.

\section{FUNDATION MODEL FOR MANAGEMENT}
The collaborative inventory foundation model refers to multiple enterprises collaborating to manage inventory by sharing information, resources, and risks from production, ordering, pricing, and recommendation. 

\subsection{Problem Formulation}

To comprehensively consider production, ordering, pricing, and recommendation, decision variables $U$, $Q$, $P$, and $K$ are defined. Managers in the supply chain can access all of this information to make coordinated decisions. Decision variables $P$ and $K$ can each influence demand to some extent, while production $U$ and ordering $Q$ affect whether demand is effectively met. The impact of $P$ and $K$ on demand follows a function:
\begin{equation*}
    D = f(P, K).
\end{equation*}
The optimization goal of the entire supply chain system is to maximize the expected profit $\Pi$ of the entire system. Profit is determined by all costs $\Pi_C$ and revenue $\Pi_\text{Rev}$ in the process. Costs consist of production costs $C_U$ and ordering costs $C_Q$. The correlation between revenue and demand effectiveness highly depends on the unit price $p_i$, inventory $n_i$, and demand $d_i$ of item $i$. The total profit of the supply chain is represented as:
\begin{equation*}
\begin{aligned}
    \Pi = \Pi_\text{Rev} - \Pi_C &=  \underset{i}{\sum} p_i \min(n_i, d_i) - C_U - C_Q,\\
    C_U &= \underset{i}{\sum} \underset{j \in J}{\sum} \mathbb{I}(u_i=j) c_\text{prod} q_i,\\
    C_Q &= \underset{i}{\sum} c_\text{order}q_i,
\end{aligned}
\end{equation*}
where $j \in J$ indicates the production method used, the production cost per unit $c_\text{prod}$ and the ordering cost $c_\text{order}$ are exogenously given.

\subsection{The Collaborative Inventory Model}

To integrate several stages including production, ordering, pricing, and recommendations, we develop an integrated collaborative decision-making model. The decision state variables encompass all known information within lead time $L$, including demand, inventory, ordering, pricing, and recommendations for all types of products. Assuming $L=1$, the state $s$ at time $t$ is represented as:
\begin{figure}
    \centering
    \includegraphics[trim=0cm 0.5cm 0cm 0cm, width = 0.93\linewidth]{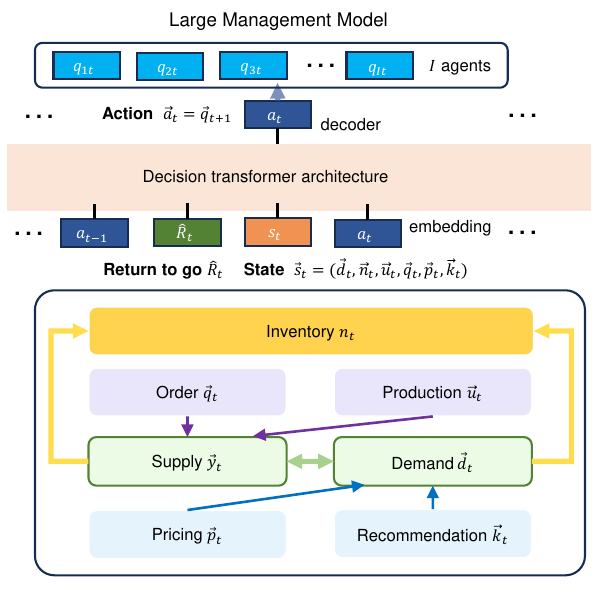}
    \caption{The overall framework of LMM based on the foundation model.}
    \label{figmodel}
\end{figure}
\begin{equation*}
    s_t = (\vec{d}_t, \vec{n}_t, \vec{u}_t, \vec{q}_t, \vec{p}_t, \vec{k}_t),
\end{equation*}
where $\vec{d}_t = (d_{1t}, d_{2t}, d_{it}, \cdots, d_{It})$, and the remaining variables $\vec{n}_t$, $\vec{u}_t$, $\vec{q}_t$, $\vec{p}_t$, $\vec{k}_t$ are similar.

The set of joint actions consisting of all decision variables is denoted by $a_t$ at time step $t$, i.e., 
\begin{equation*}
    a_t = (\vec{u}_t, \vec{q}_t, \vec{p}_t, \vec{k}_t).
\end{equation*}
The utilization of generative models in the baseline model offers an effective solution to address systematic decision-making problems. Using the transformer architecture and self-attention, an autoregressive model incorporates past trajectories of rewards, states, and actions to output joint decisions, enhancing decision-making across production, ordering, pricing, and recommendation. The trajectory is 
\begin{equation}
    \tau = (\hat{R}_1, s_1, a_1, \hat{R}_2, s_2, a_2, \cdots, \hat{R}_T, s_T, a_T),
\end{equation}
where $\hat{R}_t$ means the total reward from time step 1 to $t$, it is named return-to-go. The architecture of the LMM is shown in Fig. \ref{figmodel}. We construct the model based on Decision Transformer (DT). Prediction of future actions is achieved through linear embedding, encoding, self-attention mask, and autoregression. This structure facilitates learning intrinsic variable relations from data and achieving generalization.

\begin{figure}[h]
\centering 
\subfigure[Order decision-making.]{
\label{fig:lost_sale_imrs}
\includegraphics[trim=0cm 0.5cm 0cm 0cm, width=0.45\linewidth]{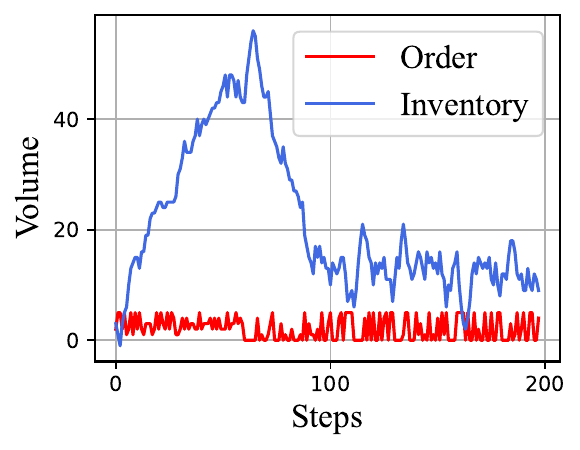}}
\subfigure[Comparison of performances.]{
\label{fig:dtact}
\includegraphics[trim=0cm 0.5cm 0cm 0cm, width=0.45\linewidth]{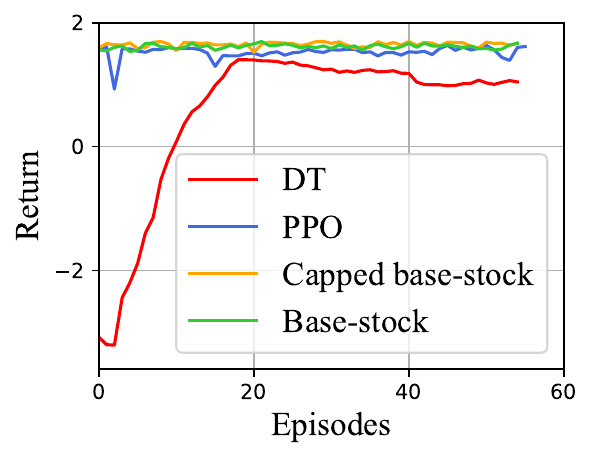}}
\caption{Performance of LMM based on the foundation model.}
\label{fig:dtret}
\end{figure}

The foundation model is used to solve the MDP. Numerical experiments in Fig. \ref{fig:dtret} demonstrate that our LMM based on the foundation model exhibits adaptability to complex scenarios, stabilizes inventory through judicious ordering decisions, and matches the performance of current state-of-the-art algorithms.

\section{CONCLUSIONS}
In this paper, we propose a general DRL framework, with the potential leading to LMM, to solve the management problems, including inventory management, dynamic pricing, and recommendation system. Our experiments illustrate the superiority of the DRL policy and the necessity of combining the three management problems. Furthermore, we propose the foundational model for LMM. The foundational model comprehensively considers inventories, pricing, and recommendations, aiming to solve the three problems with one general decision model. In the future, we will include production which is also closely related to inventory management.


\bibliographystyle{unsrt}
\bibliography{ref}

\begin{thebibliography}{10}

\bibitem{clark1960optimal}
Andrew~J Clark and Herbert Scarf.
\newblock Optimal policies for a multi-echelon inventory problem.
\newblock {\em Management science}, 6(4):475--490, 1960.

\bibitem{feng2014dynamic}
Qi~Feng, Sirong Luo, and Dan Zhang.
\newblock Dynamic inventory--pricing control under backorder: Demand estimation and policy optimization.
\newblock {\em Manufacturing \& Service Operations Management}, 16(1):149--160, 2014.

\bibitem{demirezen2016optimization}
Emre~M Demirezen and Subodha Kumar.
\newblock Optimization of recommender systems based on inventory.
\newblock {\em Production and Operations Management}, 25(4):593--608, 2016.

\bibitem{liu:2023}
Xiaotian Liu and Yijie Alexopoulos, Christos~Peng.
\newblock Deep reinforcement learning for large-scale inventory management.
\newblock {\em Available at SSRN 4490327}, 2023.

\bibitem{elmaghraby2003dynamic}
Wedad Elmaghraby and P{\i}nar Keskinocak.
\newblock Dynamic pricing in the presence of inventory considerations: Research overview, current practices, and future directions.
\newblock {\em Management science}, 49(10):1287--1309, 2003.

\bibitem{feng2020integrating}
Qi~Feng, Sirong Luo, and J~George Shanthikumar.
\newblock Integrating dynamic pricing with inventory decisions under lost sales.
\newblock {\em Management Science}, 66(5):2232--2247, 2020.

\bibitem{lei2018joint}
Yanzhe Lei, Stefanus Jasin, and Amitabh Sinha.
\newblock Joint dynamic pricing and order fulfillment for e-commerce retailers.
\newblock {\em Manufacturing \& Service Operations Management}, 20(2):269--284, 2018.

\bibitem{decisiontran}
Lili Chen, Kevin Lu, Aravind Rajeswaran, Kimin Lee, Aditya Grover, Misha Laskin, Pieter Abbeel, Aravind Srinivas, and Igor Mordatch.
\newblock Decision transformer: Reinforcement learning via sequence modeling.
\newblock {\em Advances in neural information processing systems}, 34:15084--15097, 2021.

\bibitem{trajectory}
Michael Janner, Qiyang Li, and Sergey Levine.
\newblock Offline reinforcement learning as one big sequence modeling problem.
\newblock {\em Advances in neural information processing systems}, 34:1273--1286, 2021.

\bibitem{Bommasani2021}
Rishi Bommasani, Drew~A Hudson, Ehsan Adeli, Russ Altman, Simran Arora, Sydney von Arx, Michael~S Bernstein, Jeannette Bohg, Antoine Bosselut, Emma Brunskill, et~al.
\newblock On the opportunities and risks of foundation models.
\newblock {\em arXiv preprint arXiv:2108.07258}, 2021.

\bibitem{qi2023practical}
Meng Qi, Yuanyuan Shi, Yongzhi Qi, Chenxin Ma, Rong Yuan, Di~Wu, and Zuo-Jun Shen.
\newblock A practical end-to-end inventory management model with deep learning.
\newblock {\em Management Science}, 69(2):759--773, 2023.

\bibitem{federgruen1999combined}
Awi Federgruen and Aliza Heching.
\newblock Combined pricing and inventory control under uncertainty.
\newblock {\em Operations research}, 47(3):454--475, 1999.

\bibitem{liu4395571data}
Shiyu Liu, Jun Wang, Rui Wang, Yue Zhang, Yanjie Song, and Lining Xing.
\newblock Data-driven dynamic pricing and inventory management of an omni-channel retailer in an uncertain demand environment.
\newblock {\em Available at SSRN 4395571}, 2023.

\bibitem{ricci2021recommender}
Francesco Ricci, Lior Rokach, and Bracha Shapira.
\newblock Recommender systems: Techniques, applications, and challenges.
\newblock {\em Recommender Systems Handbook}, pages 1--35, 2021.

\bibitem{ghoshal2020dilemma}
Abhijeet Ghoshal, Subodha Kumar, and Vijay Mookerjee.
\newblock Dilemma of data sharing alliance: when do competing personalizing and non-personalizing firms share data.
\newblock {\em Production and Operations Management}, 29(8):1918--1936, 2020.

\bibitem{tang2010review}
Christopher~S Tang.
\newblock A review of marketing--operations interface models: From co-existence to coordination and collaboration.
\newblock {\em International Journal of Production Economics}, 125(1):22--40, 2010.

\bibitem{Schulman:2017}
J.~Schulman, F.~Wolski, P.~Dhariwal, A.~Radford, and O.~Klimov.
\newblock Proximal policy optimization algorithms.
\newblock 2017.

\bibitem{Mnih:2016}
V~Mnih, AP~Badia, M~Mirza, A~Graves, T~Lillicrap, T~Harley, D~Silver, and K~Kavukcuoglu.
\newblock Asynchronous methods for deep reinforcement learning.
\newblock {\em Proc. Int. Conf. Mach. Learn.}, 48:1928--1937, 20--22 Jun 2016.

\bibitem{Gijsbrechts:2021}
J~Gijsbrechts, RN~Boute, JA~Van Mieghem, and DJ~Zhang.
\newblock Can deep reinforcement learning improve inventory management? {Performance} on lost sales, dual-sourcing, and multi-echelon problems.
\newblock {\em Manuf. Serv. Oper. Manag.}, 24(3), 2021.

\bibitem{snyder:2019}
Lawrence~V Snyder and Zuo-Jun~Max Shen.
\newblock {\em Fundamentals of supply chain theory}.
\newblock John Wiley \& Sons, 2019.

\bibitem{Oroojlooyjadid:2022}
A~Oroojlooyjadid, M~Nazari, LV~Snyder, and M~Tak\'{a}\v{c}.
\newblock A deep {Q}-network for the beer game: Deep reinforcement learning for inventory optimization.
\newblock {\em Manuf. Serv. Oper. Manag.}, 24(1):285--304, 2022.

\bibitem{liu:2023b}
X.~Liu, M.~Hu, Y.~Peng, and Y.~Yang.
\newblock Multi-agent deep reinforcement learning for multi-echelon inventory management.
\newblock 2023.

\bibitem{schlosser2019dynamic}
Rainer Schlosser and Keven Richly.
\newblock Dynamic pricing under competition with data-driven price anticipations and endogenous reference price effects.
\newblock {\em Journal of Revenue and Pricing Management}, 18:451--464, 2019.

\bibitem{thowsen1975dynamic}
Gunnar~T Thowsen.
\newblock A dynamic, nonstationary inventory problem for a price/quantity setting firm.
\newblock {\em Naval Research Logistics Quarterly}, 22(3):461--476, 1975.

\bibitem{chen2004coordinating}
Xin Chen and David Simchi-Levi.
\newblock Coordinating inventory control and pricing strategies with random demand and fixed ordering cost: The finite horizon case.
\newblock {\em Operations research}, 52(6):887--896, 2004.

\bibitem{bernstein2016simple}
Fernando Bernstein, Yang Li, and Kevin Shang.
\newblock A simple heuristic for joint inventory and pricing models with lead time and backorders.
\newblock {\em Management Science}, 62(8):2358--2373, 2016.

\bibitem{lilien1992marketing}
Gary~L Lilien, Philip Kotler, K~Sridhar Moorthy, et~al.
\newblock {\em Marketing models}, volume 803.
\newblock Prentice-Hall Englewood Cliffs, NJ, 1992.

\end{thebibliography}

\end{document}